\documentclass[runningheads]{llncs}
\usepackage{graphicx}
\usepackage{amsmath,amssymb}
\usepackage{url}
\usepackage{cite}
\usepackage{color}
\usepackage{wrapfig}
\usepackage{epsfig}
\usepackage{graphicx}
\usepackage{amsmath}
\usepackage{amssymb}
\usepackage{multirow}
\usepackage{booktabs}
\usepackage[table]{xcolor}
\usepackage{algorithm}
\usepackage{algorithmic}
\usepackage{arydshln}
\usepackage{threeparttable}
\usepackage{colortbl}
\usepackage{enumitem}
\usepackage{siunitx}
\usepackage{bm}
\usepackage{array}
\usepackage[table]{xcolor}
\usepackage{colortbl}
\usepackage{dblfloatfix}
\usepackage{amssymb}
\usepackage{marvosym}
\usepackage{lipsum}

\usepackage{pifont}
\newcommand{\xmark}{\ding{54}}
\definecolor{mygray}{gray}{.9}
\definecolor{ggray}{RGB}{127,127,127}
\definecolor{reda}{RGB}{192,0,0}
\definecolor{redb}{RGB}{217,148,143}
\definecolor{myyellow}{RGB}{190,144,0}
\definecolor{mygreen}{RGB}{80,100,40}
\definecolor{myblue}{RGB}{30,90,100}

\newcommand{\tabincell}[2]{\begin{tabular}{@{}#1@{}}#2\end{tabular}}

\newcommand{\ie}{\textit{i}.\textit{e}.}

\makeatletter
\newcommand{\thickhline}{
	\noalign {\ifnum 0=`}\fi \hrule height 1pt
	\futurelet \reserved@a \@xhline
}

\newcommand\blfootnote[1]{%
	\begingroup
	\renewcommand\thefootnote{}\footnote{#1}%
	\addtocounter{footnote}{-1}%
	\endgroup
}

\usepackage[pagebackref=true,breaklinks=true,letterpaper=true,colorlinks,bookmarks=false]{hyperref}

\usepackage[width=122mm,left=12mm,paperwidth=146mm,height=193mm,top=12mm,paperheight=217mm]{geometry}

\begin{document}
	
\pagestyle{headings}
\mainmatter
\def\ECCVSubNumber{1956}

\title{Weakly Supervised 3D Object Detection from \\Lidar Point Cloud}

\titlerunning{Weakly Supervised 3D Object Detection}
\authorrunning{Q. Meng, W. Wang, T. Zhou, J. Shen, L. Van Gool, D. Dai}
\author{Qinghao Meng$^{1}$\and
	\Letter Wenguan Wang$^{2}$ \and
	Tianfei Zhou$^{3}$\and\\
	Jianbing Shen$^{3}$\and
	Luc Van Gool$^{2}$\and
	Dengxin Dai$^{2}$}
\institute{$^{1}$School of Computer Science, Beijing Institute of Technology\\$^{2}$ETH Zurich~~~~$^{3}$Inception Institute of Artificial Intelligence\\
	\url{https://github.com/hlesmqh/WS3D}
}
\maketitle
\begin{abstract}
	It is laborious to manually label point cloud data for training high-quality 3D object detectors.
	This work proposes a \textit{weakly supervised} approach for 3D object detection, only requiring a small set of weakly annotated scenes, associated with a few precisely labeled object instances. This$_{\!}$ is$_{\!}$ achieved$_{\!}$ by$_{\!}$ a$_{\!}$ two-stage$_{\!}$ architecture$_{\!}$ design. Stage-1$_{\!}$ learns to generate cylindrical object proposals under weak supervision, \ie, only the horizontal centers of objects are click-annotated in bird's view scenes. Stage-2 learns to refine the cylindrical proposals to get cuboids and confidence scores, using a few well-labeled instances. Using only $500$ weakly annotated scenes and $534$ precisely labeled vehicle instances, our method achieves $85$$-$$95$\% the performance of current top-leading, fully supervised detectors (requiring $3,712$ exhaustively and precisely annotated scenes with $15,654$ instances). Moreover, with our elaborately designed network architecture, our trained model can be applied as a 3D object annotator, supporting both automatic and active (human-in-the-loop) working modes. The annotations generated by our model can be used to train 3D object detectors, achieving over $94$\% of their original performance (with manually labeled training data). Our experiments also show our model's potential in boosting performance when given more training data. Above designs make our approach highly practical and introduce new opportunities for learning 3D object detection at reduced annotation cost.
\end{abstract}

\section{Introduction}

Over the past several years\blfootnote{\Letter~Corresponding author: \textit{Wenguan Wang} (wenguanwang.ai@gmail.com).}, extensive industry and research efforts have been dedicated to autonomous driving. Significant progress has been made in key technologies for innovative autonomous driving functions, with 3D object detection being one representative example. Almost all recent successful 3D object detectors are built upon \textit{fully supervised} frameworks. They provided various solutions to problems arising from monocular images\!~\cite{Chen_2016_CVPR,Chabot_2017_CVPR}, stereo images\!~\cite{Xiaozhi2015} or point clouds\!~\cite{Chen_2017_CVPR,Yang_2018_CVPR,Chen_2019_ICCV,Lang_2019_CVPR}; gave insight into point cloud representation, introducing techniques such as voxelization\!~\cite{Maturana2015,Wu_2015_CVPR} and point-wise operation\!~\cite{Qi_2017_CVPR}; and greatly advanced the state-of-the-arts. However, these methods necessitate \textit{large-scale, precisely-annotated} 3D data to reach performance saturation and avoid overfitting. Unfortunately, such data requirement involves an astonishing amount of manual work, as it takes hundreds of hours to annotate just one hour of driving data. The end result is that a corpus of 3D training data is not only costly to create, but also limited in size and variety. In short,
the demand for massive, high-quality yet expensive labeled data has become one of the biggest challenges faced by 3D object detection system developers.

\begin{figure}[t]
	\centering
	\includegraphics[width=0.98\linewidth]{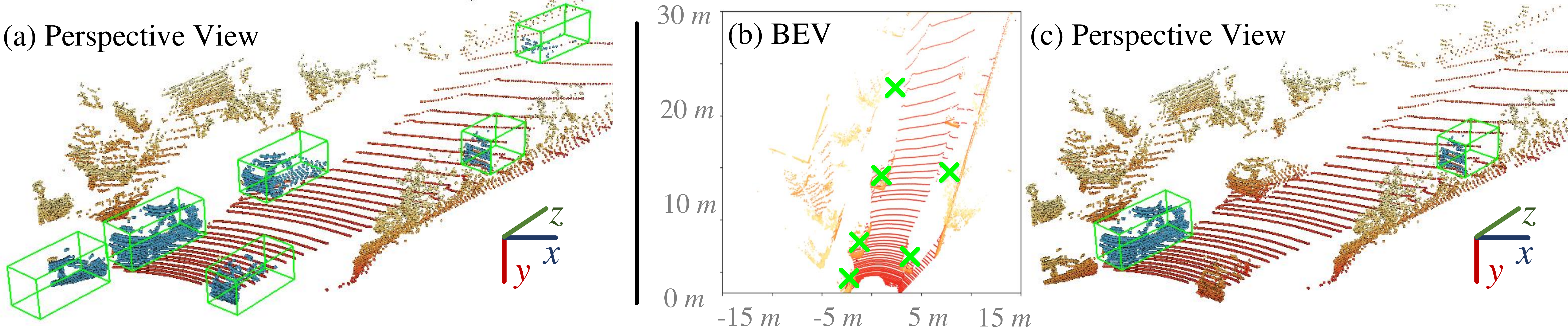}
	\caption{{Comparison between the \textit{full} supervision used in prior arts \textbf{(a)} and our \textit{inaccurate}, \textit{inexact} \textbf{(b)} and \textit{incomplete} \textbf{(c)} supervision.  Previous fully supervised methods are trained from massive, exhaustively-labeled scenes
			($3,712$ precisely annotated scenes with $15,654$ vehicle instances), while our model uses  only $500$ weakly annotated scenes with center-annotated BEV maps as well as $534$ precisely labeled vehicle instances.
	}}
	\label{fig:supervisioncom}
\end{figure}

In order to promote the deployment of 3D object detection systems, it is necessary to decrease the heavy annotation burden. However, this essential issue has not received due attention so far. To this end, we propose a weakly supervised method that learns 3D object detection from less training data, with more easily-acquired and cheaper annotations. Specifically, our model has two main stages. Stage-1 learns to predict the object centers on the $(x,z)$-plane and identity foreground points. The training in this stage only requires a small set of weakly annotated bird's eye view (BEV) maps, where the horizontal object centers are labeled (Fig.~\ref{fig:supervisioncom}(b)). Such \textit{inexact} and \textit{inaccurate} supervision greatly saves annotation efforts. Since the height information is missing in BEV maps, we generate a set of cylindrical proposals whose extent along the $y$-axis is unlimited. Then, Stage-2 learns to estimate 3D parameters from these proposals and predict corresponding confidence scores.  The learning paradigm in this stage is achieved by a few, precisely-annotated object instances as \textit{incomplete} supervision (Fig.\!~\!\ref{fig:supervisioncom}(c)), in contrast to prior {arts}\!~\cite{Yang_2018_CVPR,Li2016VehicleDF}, which consume massive, exhaustively-labeled scenes (\textit{full} ground-truth labels, Fig.~\ref{fig:supervisioncom}(a)).

Our weakly supervised framework provides two appealing characteristics.
First, it learns 3D object detection by making use of a small amount of weakly-labeled BEV data and precisely-annotated object instances. The weak supervision from BEV maps is in the form of click annotations of the horizontal object centers.
This enables much faster data labeling compared to strong supervision requiring cuboids  to be elaborately annotated on point clouds (about $40$$\sim$$50$$\times$ faster; see \S\ref{sec:3Dannotation}). For the small set of well-annotated object instances, we only label $25$\% of objects in the weakly-labeled scenes, which is about $3\%$ of the supervision used in current leading models.
Such a weakly supervised 3D object detection paradigm not only provides the opportunity to reduce the strong supervision requirement in this field, but also introduces immediate commercial benefits.

Second, once trained, our detector can be applied as an annotation tool to assist the laborious labeling process.
Current popular,  supervised solutions eagerly consume all the training data to improve the performance, while pay little attention to how to facilitate the training data annotation. Our model design allows both automatic and active working modes. In the automatic mode (no annotator in the loop), after directly applying our model to automatically re-annotate KITTI dataset\!~\cite{KITTI}, re-trained PointPillars\!~\cite{Lang_2019_CVPR} and PointRCNN\!~\cite{Shi_2019_CVPR} can maintain more than $94$\% of their original performance. In the active setting, human annotators first provide center-click supervision on the BEV maps, which is used as privileged information to guide our Stage-2 for final cuboid prediction.
Under such a setting, re-trained PointPillars and PointRCNN reach above $96$\% of their original performance. More essentially, compared with current strongly supervised annotation tools\!~\cite{zakharov2019autolabeling,lee2018leveraging}, our model is able to provide more accurate annotations with much less and weaker supervision, at higher speed (\S\ref{sec:ex-an}).

\begin{wrapfigure}[15]{r}{0.52\linewidth}
	\vspace{-8mm}
	\includegraphics[width=1\linewidth]{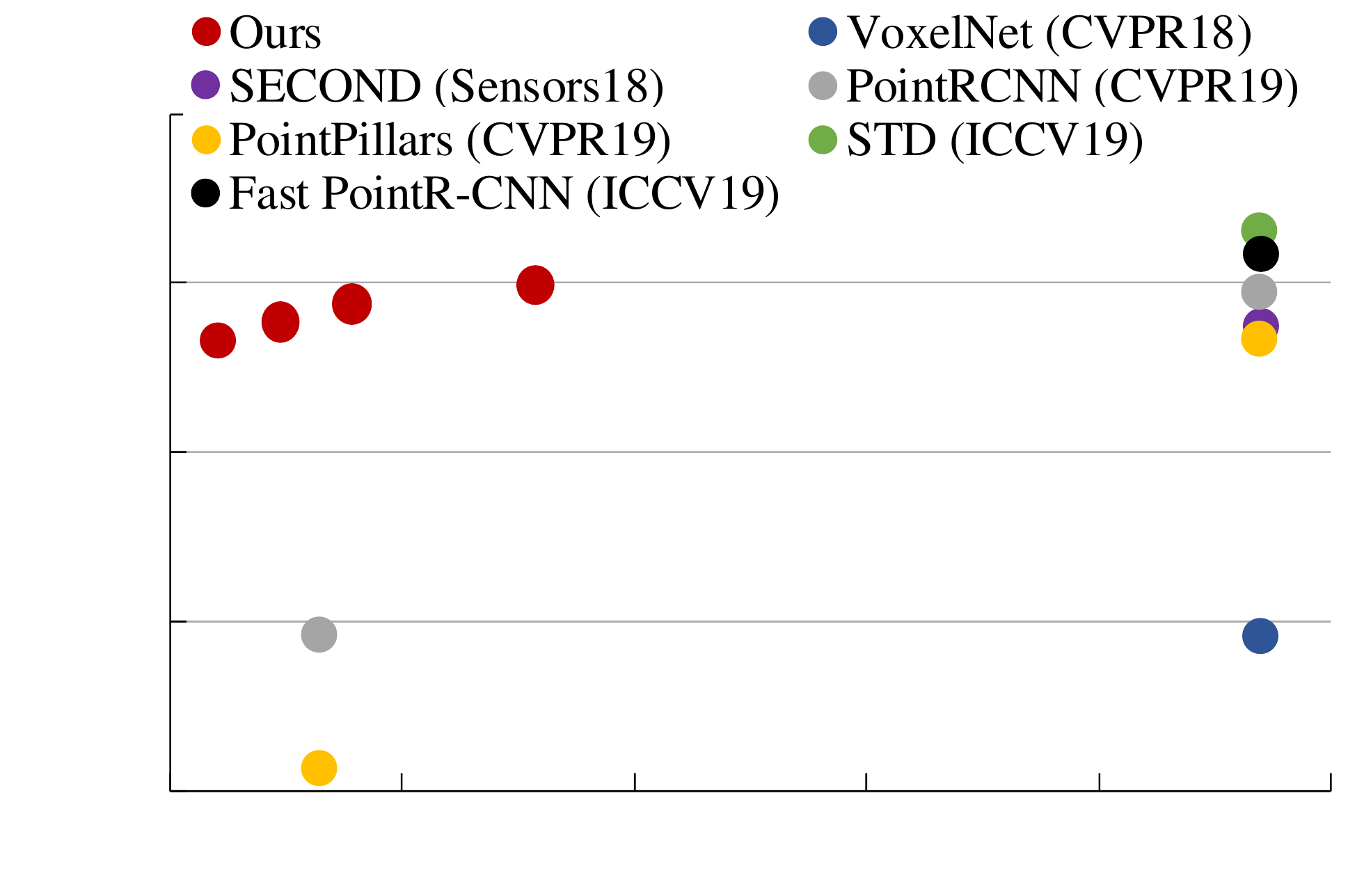}
	\put(-173,12){\fontsize{7pt}{7pt}\selectfont $60.0$}
	\put(-173,35){\fontsize{7pt}{7pt}\selectfont $66.0$}
	\put(-173,57){\fontsize{7pt}{7pt}\selectfont $72.0$}
	\put(-173,78){\fontsize{7pt}{7pt}\selectfont $78.0$}
	\put(-173,99){\fontsize{7pt}{7pt}\selectfont $84.0$}
	\put(-161,5){\fontsize{7pt}{7pt}\selectfont $0$}
	\put(-137,5){\fontsize{7pt}{7pt}\selectfont $10^5$}
	\put(-112,5){\fontsize{7pt}{7pt}\selectfont $2\!\times\!10^5$}
	\put(-80,5){\fontsize{7pt}{7pt}\selectfont $3\!\times\!10^5$}
	\put(-49,5){\fontsize{7pt}{7pt}\selectfont $4\!\times\!10^5$}
	\put(-19,5){\fontsize{7pt}{7pt}\selectfont $5\!\times\!10^5$}
	\put(-93.5,105){\fontsize{6pt}{6pt}\selectfont \cite{Yan2018}}
	\put(-92.5,98.5){\fontsize{6pt}{6pt}\selectfont \cite{Lang_2019_CVPR}}
	\put(-78.5,91.5){\fontsize{6pt}{6pt}\selectfont \cite{Chen_2019_ICCV}}
	\put(-11,105.5){\fontsize{6pt}{6pt}\selectfont \cite{Shi_2019_CVPR}}
	\put(-16.5,112.5){\fontsize{6pt}{6pt}\selectfont \cite{Zhou_2018_CVPR}}
	\put(-30.5,98.5){\fontsize{6pt}{6pt}\selectfont \cite{Yang_2019_ICCV}}
	\put(-132,-5){\fontsize{8pt}{7pt}\selectfont Annotation Cost (second)}
	\put(-184,10){\fontsize{8pt}{7pt}\selectfont  {\rotatebox{90}{Average Precision@IoU0.7}}}
	\vspace{-2mm}
	\caption{$_{\!}$Performance$_{\!}$ \textit{vs.}$_{\!}$ annotation$_{\!}$ efforts$_{\!}$, tested$_{\!}$ on$_{\!}$ KITTI\!~\cite{KITTI}$_{\!}$ \texttt{val}$_{\!}$ set (\textbf{Car}), under$_{\!}$ the$_{\!}$ \textit{moderate}$_{\!}$ regime. Our$_{\!}$ model$_{\!}$ yields$_{\!}$ promising$_{\!}$
		results$_{\!}$ with$_{\!}$ far$_{\!}$ less$_{\!}$ annotation$_{\!}$ demand (\S\ref{sec:ex-qp}$_{\!}$).$_{\!}$}
	\label{fig:annotationcom}
\end{wrapfigure}
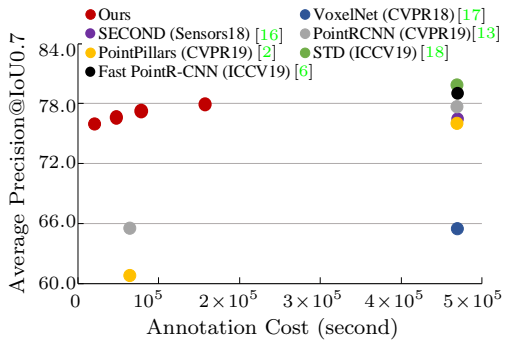

For KITTI\!~\cite{KITTI}, the experiments on \textbf{Car} class show that, using only $500$ weakly annotated scenes and $534$ precisely labeled vehicle instances, we achieve $85$$-$$95$\% of the performance of fully supervised state-of-the-arts (which require $3,712$ precisely annotated scenes with $15,654$ vehicle instances). When using more training data, our performance is further boosted (Fig.\!~\ref{fig:annotationcom}).
For \textbf{Pedestrian} class with fewer annotations, our method even outperforms most existing methods, clearly demonstrating the effectiveness of our proposed weakly supervised learning paradigm.

\section{Related Work}
\noindent\textbf{Learning Point Cloud Representations:} Processing sparse, unordered point cloud data from LiDAR sensors is a fundamental problem in many 3D related areas. There are two main paradigms for this: voxelization or point based methods. The first type of methods\!~\cite{Maturana2015,Xiang_2015_CVPR,Wu_2015_CVPR,xie2018learning} voxelize point clouds into volumetric grids and apply 2D/3D CNNs for prediction. Some of them\!~\cite{Li2016VehicleDF,Su_2015_ICCV,Qi_2016_CVPR,Chen_2017_CVPR} further improve volumetric features with multi-view representations of point clouds. Voxel based methods are computationally efficient but suffer from information loss (due to quantitization of point clouds with coarse voxel resolution).
The second-type, point based methods\!~\cite{Qi_2017_CVPR,rethage2018eccv,qi2017,Qi_2018_CVPR,Zhou_2018_CVPR,Shi_2019_CVPR}, which directly operate on raw point clouds and preserve the original information, recently became popular.

\noindent\textbf{3D Object Detection:} A flurry of techniques have been
explored for 3D object detection in driving scenarios, which can be broadly categorized into three classes.  %
\textit{(1)$_{\!}$ 2D$_{\!}$ image$_{\!}$ based$_{\!}$ methods}$_{\!}$ focus$_{\!}$  on$_{\!}$  camera$_{\!}$  based$_{\!}$  solutions$_{\!}$  with$_{\!}$  monocular$_{\!}$  or$_{\!}$  stereo$_{\!}$  images, by exploring geometries between 3D and
2D bounding boxes\!~\cite{Xiaozhi2015,Chen_2016_CVPR,Mousavian_2017_CVPR,Li_2019_CVPR}, or similarities between 3D objects and CAD models\!~\cite{Chabot_2017_CVPR,Mottaghi_2015_CVPR}. Though efficient, they struggle against the inherent difficulty of directly estimating depth information from images. \textit{(2) 3D point cloud based methods} rely on depth sensors
such as LiDAR. Some representative ones project point clouds to bird's view
and use 2D CNNs\!~\cite{Chen_2017_CVPR,Yang_2018_CVPR} to learn the point cloud  features. Some others\!~\cite{Su_2015_ICCV,Zhou_2018_CVPR} apply 3D CNNs over point cloud voxels to generate cuboids. They tend to capture local information, due to the limited receptive fields of CNN kernels. Thus sparse convolutions\!~\cite{Yan2018} with enlarged receptive fields are adopted later.  To avoid losing information during voxelization, some efforts learn point-wise features directly from raw point clouds, using PointNet\!~\cite{Qi_2017_CVPR}-like structures\!~\cite{Chen_2019_ICCV,Lang_2019_CVPR}. \textit{(3) Fusion-based methods} \cite{Qi_2017_CVPR,qi2017,Qi_2018_CVPR,Liang_2018_ECCV,Liang_2019_CVPR,Shi_2019_CVPR} attempt to fuse information from different sensors, such as cameras and LiDAR. The basic idea is to leverage the complementary information of camera images and point clouds, \ie,
rich visual information of images and precise depth details of point clouds, to improve the detection accuracy.  However, fusion-based methods typically run slowly due to the need of processing multi-modal inputs\!~\cite{Yan2018}.

\noindent\textbf{Click Supervision:} Click annotation schemes were used to reduce the burden of collecting segmentation/bounding box annotations at a large scale. Current efforts typically leverage center-click\!~\cite{bearman2016s,papadopoulos2017training},  extreme-point\!~\cite{maninis2018deep,papadopoulos2017extreme}, or corrective-click\!~\cite{bearman2016s} supervision for semantic segmentation\!~\cite{benenson2019large,bearman2016s} or object detection\!~\cite{papadopoulos2017training,maninis2018deep,papadopoulos2017extreme} in 2D visual scenarios. However, in this work, we explore center clicks, located on BEV maps, as weak supervision signals for 3D object detection.

\noindent\textbf{3D Object Annotation:} Very few attempts were made to scale up 3D object annotation pipelines\!~\cite{lee2018leveraging,zakharov2019autolabeling}.  \cite{lee2018leveraging} lets an annotator place 2D seeds from which to infer 3D segments and centroid parameters, using fully supervised learning. \cite{zakharov2019autolabeling} suggests a differentiable template matching model with curriculum learning.  In addition to different annotation paradigms, model designs and level of human interventions, our model is also unique in its weakly supervised learning strategy and dual-work mode, and achieves stronger performance.

\section{Data Annotation Strategy for Our Weak Supervision}
\label{sec:3Dannotation}

Before$_{\!}$ detailing$_{\!}$ our$_{\!}$ model, we$_{\!}$ first$_{\!}$ discuss$_{\!}$ how$_{\!}$ to$_{\!}$ get$_{\!}$ our$_{\!}$ weakly$_{\!}$ supervised$_{\!}$ data.$_{\!}$

\begin{figure}[t]
	\centering
	\includegraphics[width=0.98\linewidth]{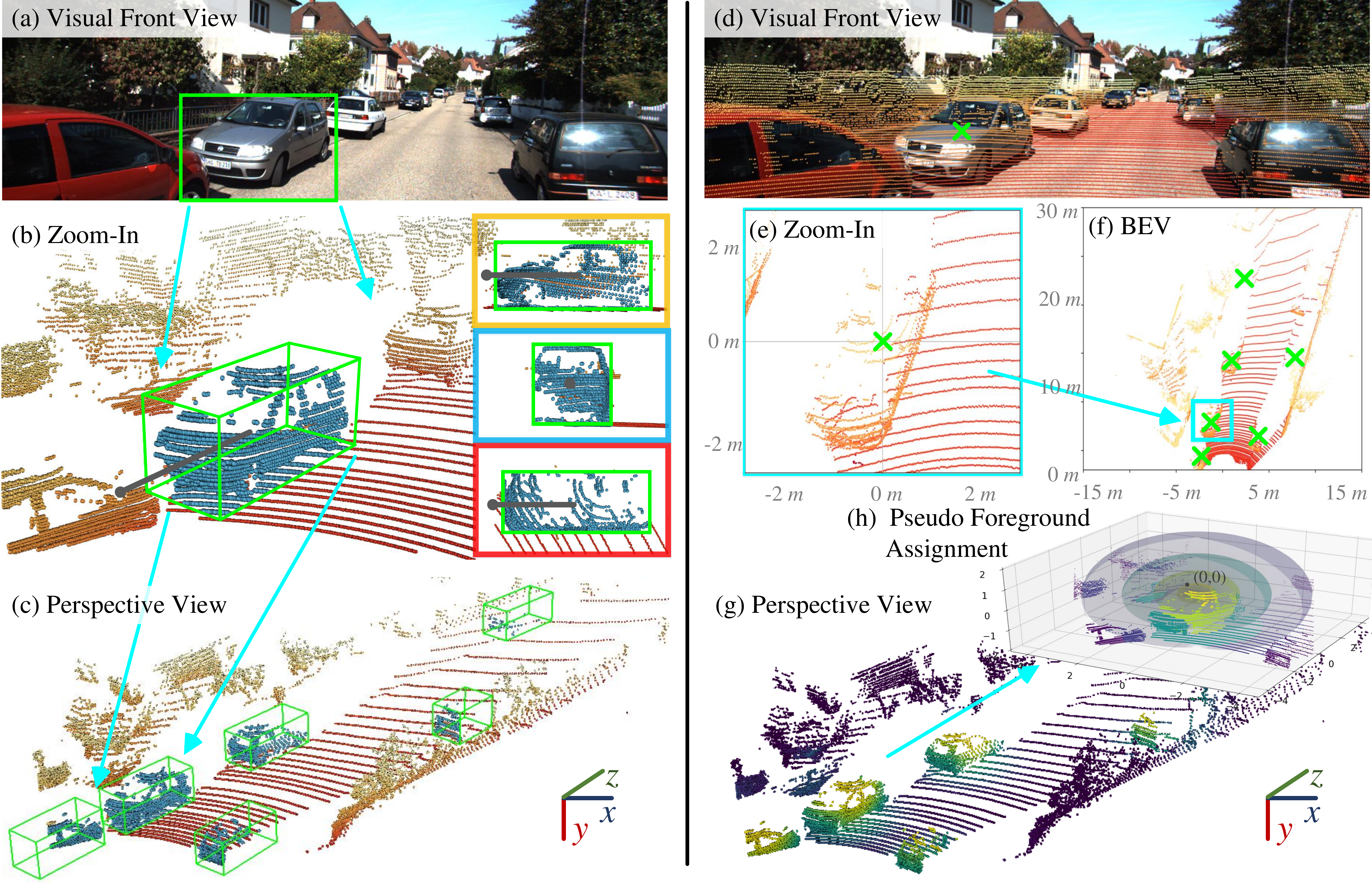}
	\put(-89,81){\fontsize{6pt}{6.2pt}\selectfont (Eq.\!~\ref{eq:Pseudo})}
	\caption{{\textbf{(a-c)}: Precise annotations require extensive labeling efforts (see \S\ref{sec:3Dannotation}). \textbf{(d-f)}: Our weak supervision is simply obtained by clicking object centers (denoted by {\color{green}\xmark}) on BEV maps (see \S\ref{sec:3Dannotation}). \textbf{(g-h)}: Our pseudo groundtruths for fore-/background segmentation (yellower indicates higher foreground score; see \S\ref{sec:stage-1}).
	}}
	\label{fig:label}
\end{figure}

\noindent\textbf{Traditional Precise But Laborious Labeling Strategy:}  Current popular 3D object detectors are fully supervised deep learning models, requiring precisely annotated data. However, creating a high-quality 3D object detection dataset is more complex than creating, for example, a 2D object detection dataset.
For precise labeling \cite{xie2016semantic,wang2019apolloscape}, annotators first navigate the 3D scene to find an object with the help of visual content from the camera image (Fig.\!~\ref{fig:label}\!~(a)). Later, an initial rough cuboid and orientation arrow (Fig.\!~\ref{fig:label}\!~(b)) are placed. Finally, the optimal annotation (Fig.\!~\ref{fig:label}\!~(c)) is obtained by gradually adjusting the 2D boxes projected in orthographic views. As can be seen, although this labeling procedure generates high-quality annotations, it contains several subtasks with gradual corrections and 2D-3D view switches. It is thus quite laborious and expensive.

\noindent\textbf{Our Weak But Fast Annotation Scheme:} Our model is learned from a small set of weakly annotated BEV maps, combined with a few precisely labeled 3D object instances. The weakly annotated data only contains object center-annotated BEV maps, which can be easily obtained. Specifically, human annotators first roughly click a target on the camera front-view map (Fig.\!~\ref{fig:label}\!~(d)). Then the BEV map is zoomed in and the region around the initial click is presented for a more accurate center-click (Fig.\!~\ref{fig:label}\!~(e)). Since our annotation procedure does not refer to any 3D view, it is very easy and fast; most annotations can be finished by only two clicks. However, the collected supervision is weak, as only the object centers over ($x$, $z$)-plane are labeled, without height information in y-axis and size of cuboids.

\noindent\textbf{Annotation Speed:} We re-labeled KITTI \texttt{train} set\!~\cite{KITTI}, which has $3,712$ driving scenes with more than 15K vehicle instances. This took about $11$ hours, \ie, $2.5$ s per instance. As KITTI does not report the annotation time, we refer to other published statistics\!~\cite{Song_2015_CVPR,wang2019apolloscape}, which suggest around $114$ s per instance in a fully manual manner\!~\cite{Song_2015_CVPR} or $30$s with extra assistance of a 3D object detector\!~\cite{wang2019apolloscape}. Thus our click supervision provides a $15$$\sim$$45$$\times$ reduction in the time required for traditional precise annotations.

\begin{wrapfigure}[9]{r}{0.5\linewidth}
	\vspace{-8mm}
	\includegraphics[width=1\linewidth]{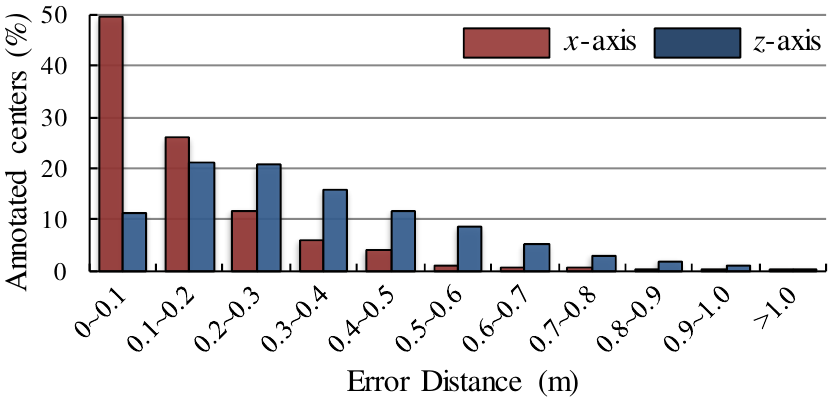}
	\vspace{-6mm}
	\caption{{Distance distributions (\textbf{Car}) on $x$- and $z$-axes of our weak BEV annotations.}}
	\label{fig:statistics}
\end{wrapfigure}

\noindent\textbf{Annotation Quality:} To assess our annotation quality, Fig.\!~\ref{fig:statistics} depicts the average  distance between our annotated centers and the KITTI groundtruths on BEV maps.
The average errors on $x$- and $z$-axes are about 0.25 m and 0.75 m, respectively, bringing out the limitation of LiDAR sensors in capturing the object better to its side than the back.

\section{Proposed Algorithm}
\label{sec:method}

Our object detector takes raw point clouds as input and outputs oriented 3D boxes. It has a cylindrical 3D proposal generation stage (\S\ref{sec:stage-1}, Fig.\!~\ref{fig:overview}(a-b)), learning from click supervision, and a
subsequent, proposal-based 3D object localization stage, learning from a few, well-annotated object instances  (\S\ref{sec:stage-2}, Fig.\!~\ref{fig:overview}(c-d)). Below we will focus on \textbf{Car} class. However, as evidenced in our experiments (\S\ref{sec:ex-qp}),  our model can also easily be applied to other classes, such as \textbf{Pedestrian}.

\subsection{\!\!Learn to Generate Cylindrical Proposals from Click Annotations}

\label{sec:stage-1}
There are two goals in our first stage: 1) to generate foreground point segmentation; and 2) to produce a set of cylinder-shaped 3D object proposals. The fore-/background separation is helpful for the proposal generation and provides useful information for the second stage. Because only the horizontal centers of objects are labeled on the BEV maps, our proposals are cylinder-shaped.

\noindent\textbf{Pseudo Groundtruth Generation.} Since the annotations in the BEV maps are weak, proper modifications should be made to produce pseudo, yet stronger supervision signals. Specifically, for a labeled vehicle center point $o\!\in\!\mathcal{O}$, its horizontal location $(x_o,z_o)$ in the LiDAR coordinate system can be inferred according to the projection from BEV to point cloud. We set its height $y_o$ (over $y$-axis) to the LiDAR sensor's height (the height of the ego-vehicle), \ie, $y_o\!=\!0$. The rationale behind such a setting will be detailed later. Then, for each unlabeled point $p$, its pseudo foreground value $f^p\!\in[0,1]\!$ is defined as:
\begin{equation}
\begin{aligned}
\!\!\!f^{p} \!=\!\max\nolimits_{o\in \mathcal{O}}(\iota(p,o)),
~~\text{where~~}\iota(p,o) \!=\!
\left\{
\begin{aligned}
&1~~~~~~~~~~~~~~~~~~\text{if~~} d(p,o)\leq0.7,\!\\
&\frac{1}{\kappa}\mathcal{N}(d(p,o))\!~~~~\text{if~~} d(p,o)>0.7.\!
\end{aligned}
\right.
\end{aligned}
\label{eq:Pseudo}
\end{equation}
Here, $\mathcal{N}$ is a 1D Gaussian distribution with mean 0.7 and variance 1.5, and $\kappa\!=\!\mathcal{N}(0.7)$ is a normalization factor. And $d(p,o)$ is a distance function: $d(p,o)\!=\![(x_p\!-\!x_o)^2\!+\!\frac{1}{2}(y_p-y_o)^2+(z_p\!-\!z_o)^2]^{\frac{1}{2}}$, where $(x_p,y_p,z_p)$ is the 3D coordinate of $p$. The coefficient ($=\!\frac{1}{2}$) is used to due to the large uncertainty over $y$-axis. The foreground probability assignment function $\iota(p,o)$ gives high confidence ($=$1) for those points close to $o$ (\ie, $d(p,o)\!\leq\!0.7$), and attenuates the confidence for distant ones  (\ie, $d(p,o)\!>\!0.7$) by following the Gaussian distribution $\mathcal{N}$.
The reason why we set the heights of the labeled center points $\mathcal{O}$ as 0 is because most object points are at lower altitudes  than the LiDAR sensor (at the top of the ego-vehicle) and they will gain high foreground values in this way.  For those background points even with similar altitudes to the LiDAR sensor, they are very sparse and typically far away from the vehicle centers in ($x$, $z$)-plane (see Fig.\!~\ref{fig:label}(h)), and$_{\!}$ can$_{\!}$ thus$_{\!}$ be$_{\!}$ ignored.  Plane detection\!~\cite{Xiaozhi2015} can be used for more accurate height estimation, but in practice we find our strategy is good enough.

\begin{figure}[t]
	\centering
	\includegraphics[width=0.99\linewidth]{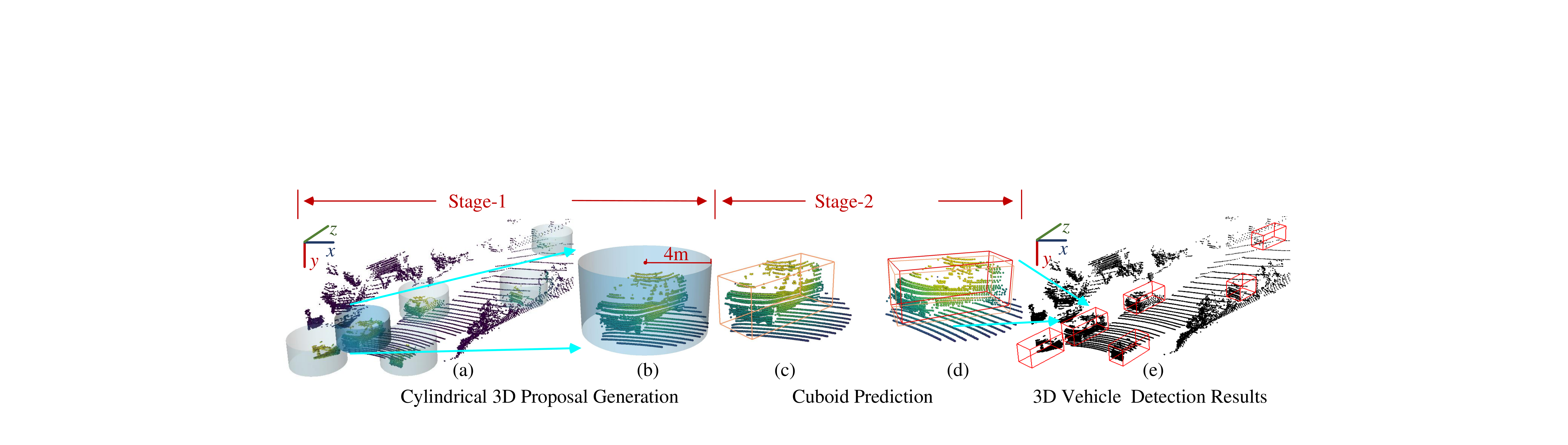}
	\put(-267.5,69){\fontsize{6pt}{6pt}\selectfont  {(\S\ref{sec:stage-1})}}
	\put(-142,69){\fontsize{6pt}{6pt}\selectfont  {(\S\ref{sec:stage-2})}}
	\caption{{Our 3D object detection pipeline (\S\ref{sec:method}). (a-b) Cylindrical 3D proposal generation results from Stage-1 (\S\ref{sec:stage-1}). Yellower colors correspond to higher foreground probabilities.
			(c-d) Cuboid prediction in Stage-2 (\S\ref{sec:stage-2}).  (e) Our final results.
		}
	}
	\label{fig:overview}
\end{figure}

\noindent\textbf{Point Cloud Representation.} Several set-abstraction
layers with multi-scale grouping are applied to directly learn discriminative point-wise features from raw point cloud input\!~\cite{qi2017}. Then, two branches are placed over the backbone for foreground point segmentation and vehicle ($x,z$)-center prediction, respectively.

\noindent\textbf{Foreground Point Segmentation.}
With the point-wise features extracted from the backbone network and pseudo groundtruth $f^{p}$ generated in Eq.~\ref{eq:Pseudo}, the foreground segmentation branch learns to estimate the foreground probability $\tilde{f}^{p}$ of each point $p$. The learning is achieved by minimizing the following loss:\!\!\!
\begin{equation}
\begin{aligned}
\mathcal{L}_{\text{seg}}=\alpha(1-\hat{{f}}^{p})^\gamma\log(\hat{{f}}^{p}), \text{~~~~where~~}\hat{{f}}^{p}=\tilde{f}^{p}\cdot f^{p}+(1-\tilde{f}^{p})\cdot(1- f^{p}).
\end{aligned}
\label{eq:loss1}
\end{equation}
This is a soft version of the focal loss\!~\cite{Lin_2017_ICCV}, where the binary fore-/background label is given in a probability formation. As in\!~\cite{Lin_2017_ICCV}, we set $\alpha\!=\!0.25$ and $\gamma\!=\!2$.

\noindent\textbf{Object ($x,z$)-Center Prediction.} The other branch is for object ($x,z$)-center regression, as the weakly annotated BEV maps only contain horizontal information. As in\!~\cite{Shi_2019_CVPR}, a bin-based classification strategy is adopted. For each labeled object center $o\!\in\!\mathcal{O}$, we set the points within 4 m distance as \textit{support points} (whose pseudo foreground probabilities $\!\geq\!0.1$). These support points are used to estimate the horizontal coordinates of $o$. For each support point $p$, its surrounding area ($L\!\times\!L$ m$^2$) along $x$- and $z$- axes is the searching space for $o$, which is split into a series of discrete bins. Concretely, for $x$- and $z$- axis, the search range $L (=_{\!}\!8\!~\text{m})$ is divided into 10 bins of uniform length $\delta (=_{\!}\!0.8\!~\text{m})$.
Therefore, for a support point $p$ and the corresponding center $o$, the target bin assignments ($b_x, b_z$) along $x$- and $z$-axis can be formulated as:
\begin{equation}
\begin{aligned}
b_x=\lfloor\frac{x_p-x_o+L}{\delta}\rfloor, ~~~~b_z=\lfloor\frac{z_p-z_o+L}{\delta}\rfloor.
\end{aligned}
\label{eq:bin}
\end{equation}
Residual ($r_x, r_z$) is computed for further location refinement within each assigned bin:
\begin{equation}
\begin{aligned}
r_{u\in\{x,z\}}=\frac{u}{\varepsilon}(u_p-u_o+L-(b_u\cdot\delta+\frac{\delta}{2})), \text{~~~~where~~} \varepsilon\!=\!\frac{\delta}{2}.
\end{aligned}
\label{eq:residual}
\end{equation}
For a support point $p$, the center localization loss $\mathcal{L}_{\text{bin}}$ is designed as:
\begin{equation}
\begin{aligned}
\mathcal{L}_{\text{bin}}=\sum\nolimits_{u\in\{x,z\}}\mathcal{L}_{\text{cls}}(\tilde{b}_u, b_u)+\mathcal{L}_{\text{reg}}(\tilde{r}_u, r_u),\\
\end{aligned}
\label{eq:loss2}
\end{equation}
where $\tilde{b}$ and $\tilde{r}$ are predicted bin assignments and residuals, and $b$ and $r$ are the
targets.  $\mathcal{L}_{\text{cls}}$ is a cross-entropy
loss for bin classification along the ($x,z$)-plane, and $\mathcal{L}_{\text{reg}}$ refers to the $\ell$1 loss for residual regression w.r.t the target bins.

\noindent\textbf{Cylindrical 3D Proposal Generation.} During inference, the segmentation branch estimates the foreground probability of each point. Then, we only preserve the points whose foreground scores are larger than 0.1.
As we only have horizontal coordinates of the centers, we cannot directly generate 3D bounding box proposals. Instead, for each center, we generate a cylindrical proposal with a 4 m radius over ($x,z$)-plane and unlimited extent along $y$-axis (Fig.\!~\ref{fig:overview}\!~(a, b)).

\noindent\textbf{Center-Aware Non-Maximum Suppression.} To eliminate redundant proposals, we propose a center-aware non-maximum suppression (CA-NMS) strategy. The main idea is that, {it is easier to predict centers from center-close points than far ones, and center-close points gain
	high foreground scores under our pseudo groundtruth generation strategy.} Thus, for a predicted center, we use the foreground probability of its sourced (support) point, as its confidence score. That means we assume that a point with a higher foreground score is more center-close and tends to make a more confident center prediction. Then  we rank all the predicted centers according to their confidence, from large to small. For each center, if its distance to any other pre-selected centers is larger than 4~m on the ($x,z$)-plane,  its proposal will be preserved; otherwise it is removed.

\subsection{Learn to Refine Proposals from A Few Well-Labeled Instances}
\label{sec:stage-2}
Stage-2 is to estimate cuboids from proposals and recognize false estimates. We achieve this by learning a proposal refinement model from a few well-annotated instances, motivated by two considerations.  \textbf{(i)} The proposal refinement is performed instance-wise, driving us to consume instance-wise annotations. \textbf{(ii)} Our initial cylindrical proposals, though rough, contain rich useful information, which facilitates cuboid prediction especially when training data is limited.

\noindent\textbf{Overall Pipeline.} Our method carries out refinement of cuboid predictions over two steps. First, an initial cuboid generation network takes cylindrical proposals as inputs, and outputs initial cuboid estimations (Fig.\!~\ref{fig:overview}\!~(b-c)). Then, a final cuboid refinement network takes the initial cuboid estimations as inputs, and outputs final cuboid predictions (Fig.\!~\ref{fig:overview}\!~(c-d)) as well as confidence.

\noindent\textbf{Initial Cuboid Generation.} The initial cuboid generation network  stacks several set abstraction layers, intermediated with a single-scale grouping operation, to collect contextual and pooled point features as the cylindrical proposal representations\!~\cite{qi2017}. Then, a multilayer perceptron based branch is appended for initial cuboid estimation. Let us denote the groundtruth of an input cuboid as $(x,y,z,h,w,l,\theta)$, where $(x,y,z)$ are the object-center coordinates, $(h,w,l)$ object size, and $\theta$ orientation from BEV. A bin-based regression loss $\mathcal{L}_{\text{bin}}$ is applied for estimating $\theta$, and a smooth $\ell$1 loss $\mathcal{L}_{\text{reg}}$ is used for other parameters:
\begin{equation}
\begin{aligned}
\mathcal{L}_{\text{ref}}=\mathcal{L}_{\text{bin}}(\tilde{\theta}, \theta)+\sum\nolimits_{u\in\{x,y,z,h,w,l\}}\mathcal{L}_{\text{reg}}(\tilde{u}, u),
\end{aligned}
\label{eq:loss3-1}
\end{equation}
where $(\tilde{x},\tilde{y},\tilde{z},\tilde{h},\tilde{w},\tilde{l},\tilde{\theta})$ are the estimated cuboid parameters.

\noindent\textbf{Final Cuboid Refinement.} The final cuboid refinement network has the similar network architecture of the initial cuboid generation network. It learns to refine initial cuboid estimations with the same loss design in Eq.~\ref{eq:loss3-1}. In addition, to predict cuboid's confidence, an extra confidence estimation head is added, which is supervised by an IoU-based regression loss\!~\cite{Li_2019_CVPR,Shi_2019_part}:
\begin{equation}
\begin{aligned}
~~~~\mathcal{L}_{\text{con}}=\mathcal{L}_{\text{reg}}(\tilde{C}_{\text{IoU}}, C_{\text{IoU}}),
\end{aligned}
\label{eq:loss3-2}
\end{equation}
where the targeted confidence score $C_{\text{IoU}}$ is computed as the largest IoU score between the output cuboid and groundtruths.

In the first cuboid generation step, for each groundtruth 3D bounding box, cylindrical proposals whose center-distances (on ($x,z$)-plane) are less than 1.4 m away are selected as the training samples. Then, the output cuboids from those cylindrical proposals are further used as the training samples for the groundtruth in the final refinement step.

\subsection{Implementation Detail}
\label{sec:implementation}

\noindent\textbf{Detailed Network Architecture.} In Stage-1 (\S\ref{sec:stage-1}), to align the network input, 16K points are
sampled from each point-cloud scene. Four set-abstraction layers with multi-scales are stacked to sample the points into groups with sizes ($4096$, $1024$, $256$, $64$).
Four feature propagation layers are then used to obtain point-wise features, as the input for the segmentation and center prediction branches. The segmentation branch contains two FC layers with $128$ and $1$ neuron(s), respectively. The ($x,z$)-center prediction branch has two FC layers with $128$ and $40$ neurons, respectively. In Stage-2 (\S\ref{sec:stage-2}), $512$ points are sampled from each cylindrical proposal/cuboid, and each point is associated with a 5D feature vector, \ie, a concatenation of 3D point coordinates, 1D laser reflection intensity, and foreground score. Before feeding each proposal/cuboid into the generation/refinement network, the coordinates of points are canonized to guarantee their translation and rotation invariance\!~\cite{Chen_2019_ICCV}. The corresponding groundtruth is modified accordingly. In addition, for cylindrical proposals, only a translation transformation is performed over $(x,z)$-plane, \ie, the horizontal coordinates of the proposal center are set as $(0,0)$. For each cuboid, the coordinates of points within a 0.3 m radius are cropped to include more context. For the cuboid generation/refinement network, four set-abstraction layers with single-scale grouping are used to sample the input points into groups with sizes ($256$, $128$, $32$, $1$). Finally, a 512D feature is extracted for cuboid and confidence estimation.

\noindent\textbf{Data Preparation.} KITTI \texttt{train} set has $3,712$ precisely annotated training scenes with $15,654$ vehicle instances.  Unless otherwise noted, we use the following training data setting. The first $500$ scenes with our weakly annotated BEV maps are used for training our Stage-1 model, and 25\% of the vehicle instances ($=\!534$) in the $500$ scenes are associated with precise 3D annotations and used for training our Stage-2 model\footnote{The instances are randomly selected and the list will be released.}. We make use of this weak and limited training data to better illustrate the advantage of our model. This also allows us to investigate the performance when using our model as an annotation tool (see \S\ref{sec:ex-an}).

\noindent\textbf{Data Augmentation.} During training, we adopt several data augmentation techniques to avoid overfitting and improve the generalization ability. In Stage-1, left-right flipping, scaling from [0.95,  1.05], and rotation from [-10$^\circ$, 10$^\circ$] are randomly applied for each scene.  In addition, to diversify training scenarios, we randomly sample a few annotated vehicle centers with surrounding points within a cylinder with a 4~m radius, and insert them into the current sample. Furthermore, to increase the robustness to  distant vehicles, which typically contain very few points, we randomly drop the points within the cylindrical space (with a 4~m radius) of labeled centers. In Stage-2, for each proposal, we randomly conduct left-right flipping, scaling from [0.8, 1.2], and rotation from [-90$^\circ$, 90$^\circ$]. We shift each proposal by small translations, following a Gaussian distribution with mean 0 and variance 0.1, for $x$-, $y$-, and $z$-axis each individually. We randomly change the foreground label of the points. To address large occlusions, we randomly omit part of a proposal (1/4$-$3/4 of the area in BEV). Finally,
for each proposal, we randomly remove the inside points (at least 32 points remain).

\noindent\textbf{Inference.} After applying CA-NMS for the cylindrical proposals generated in Stage-1 (\S\ref{sec:stage-1}), we feed the remaining ones to  Stage-2 (\S\ref{sec:stage-2}) and get final 3D predictions. We then use an oriented NMS with a BEV IoU threshold of 0.3 to reduce redundancy. Our model runs at about 0.2 s per scene, which is on par with MV3D\!~\cite{Chen_2017_CVPR} (0.36 s), VoxelNet\!~\cite{Zhou_2018_CVPR} (0.23 s) and F-PointNet\!~\cite{Qi_2018_CVPR} (0.17 s).

\section{Experiment}
\label{sec:exp}

\subsection{Experimental Setup}

\noindent\textbf{Dataset.} Experiments are conducted on KITTI\!~\cite{KITTI}, which contains $7,481$ images
for train/val and $7,518$ images for testing. The \texttt{train/val} set has 3D bounding box groundtruths and is split into two sub-sets\!~\cite{Chen_2017_CVPR,Zhou_2018_CVPR}: \texttt{train} ($3,712$ images) and \texttt{val} ($3,769$ images). We train our detector only on a weakly labeled subset of \texttt{train} set, while the \texttt{val} set is used for evaluation only.
Detection outcomes are evaluated in the three standard regimes: \textit{easy, moderate, hard}.

\noindent\textbf{Evaluation$_{\!}$ Metric.} Following\!~\cite{Chen_2017_CVPR}, average$_{\!}$ precisions$_{\!}$ for$_{\!}$ BEV$_{\!}$ and$_{\!}$ 3D$_{\!}$ boxes$_{\!}$ are$_{\!}$ reported. Unless$_{\!}$ specified, the$_{\!}$ performance$_{\!}$ is$_{\!}$ evaluated$_{\!}$ with$_{\!}$ a$_{\!}$ 0.7 IoU$_{\!}$ threshold$_{\!}$.

\begin{table}[t]
	\caption{{\textbf{Evaluation results on KITTI \texttt{val} set (Car).} See \S\ref{sec:ex-qp} for details.}}
	\centering\small
	\label{table:KITTIval}
	\resizebox{0.99\textwidth}{!}{
		\setlength\tabcolsep{2.5pt}
		\renewcommand\arraystretch{1.00}
		\begin{tabular}{c|r||c|ccc|ccc}
			\hline\thickhline
			\rowcolor{mygray}
			&	 &  	 & \multicolumn{3}{c|}{BEV@0.7} &  \multicolumn{3}{c}{3D Box@0.7} \\ \cline{4-9}
			\rowcolor{mygray}
			\multirow{-2}{*}{Learning Paradigm} &\multirow{-2}{*}{Detector~~~~}	& \multirow{-2}{*}{Modality}&Easy &Moderate &Hard &Easy &Moderate &Hard \\  \hline \hline
			\multicolumn{9}{c}{\footnotesize{Trained with the whole KITTI \texttt{train} set:} \footnotesize{{\color{red}$3,712$ precisely labeled scenes with $15,654$ vehicle instances}}}\\
			\hline
			\multirow{8}{*}{\textit{Fully supervised}}
			&VeloFCN\!~\cite{Li2016VehicleDF} &LiDAR&40.14 &32.08 &30.47  &15.20 &13.66 &15.98\\
			&PIXOR\!~\cite{Yang_2018_CVPR}&LiDAR&86.79&80.75&76.60&-&-&-\\
			&VoxelNet\!~\cite{Zhou_2018_CVPR} &LiDAR &89.60 &84.81 &78.57&81.97 &65.46 &62.85\\
			&SECOND\!~\cite{Yan2018}&LiDAR &89.96&87.07&79.66&87.43&76.48&69.10\\
			&PointRCNN\!~\cite{Shi_2019_CVPR} &LiDAR &-&-&-&88.45&77.67&76.30\\
			&PointPillars\!~\cite{Lang_2019_CVPR}&LiDAR &89.64&86.46&84.22&85.31&76.07&69.76\\
			&Fast PointR-CNN\!~\cite{Chen_2019_ICCV}&LiDAR &90.12 &88.10 &86.24&89.12 &79.00 &77.48\\
			&STD\!~\cite{Yang_2019_ICCV}&LiDAR &90.50&88.50&88.10&89.70&79.80&79.30\\
			\hline
			\hline
			\multicolumn{9}{c}{\footnotesize{Trained with a part of KITTI \texttt{train} set:} {\footnotesize{\color{red}$500$ precisely labeled scenes with $2,176$ vehicle instances}}}\\
			\hline
			\multirow{2}{*}{\textit{Fully supervised}}&PointRCNN\!~\cite{Shi_2019_CVPR}&LiDAR &87.21&77.10&76.63&79.88&65.50&64.93\\
			&PointPillars\!~\cite{Lang_2019_CVPR}&LiDAR &86.27&77.13&75.91&72.36&60.75&55.88\\
			\hline
			\hline
			\multicolumn{9}{c}{\footnotesize{Trained with a part of KITTI \texttt{train} set:} \footnotesize{{\color{red}$125$ precisely labeled scenes with $550$ vehicle instances}}}\\
			\hline
			\multirow{2}{*}{\textit{Fully supervised}}&PointRCNN\!~\cite{Shi_2019_CVPR}&LiDAR &85.09&74.35&67.68&67.54&54.91&51.96\\
			&PointPillars\!~\cite{Lang_2019_CVPR}&LiDAR &85.76&75.30&73.29&65.51&51.45&45.53\\
			\hline \hline
			\multicolumn{9}{c}{\footnotesize{Trained with a part of KITTI \texttt{train} set:} \footnotesize{{\color{red}$500$ weakly labeled scenes with $534$ precisely annotated instances}}}\\
			\hline
			\textit{Weakly supervised}&\textbf{Ours}&LiDAR
			&88.56  &84.99  &84.74  &84.04  &75.10  &73.29\\
			\hline	
		\end{tabular}
	}
\end{table}

\begin{table}[!t]
	\caption{{\textbf{Evaluation results on KITTI \texttt{test} set (Car).} See \S\ref{sec:ex-qp} for details.}}
	\label{table:KITTItest}
	\centering\small
	\resizebox{0.99\textwidth}{!}{
		\setlength\tabcolsep{2.5pt}
		\renewcommand\arraystretch{1.0}
		\begin{tabular}{c|r||c|ccc|ccc}
			\hline\thickhline
			\rowcolor{mygray}
			&  	 && \multicolumn{3}{c|}{BEV@0.7} &  \multicolumn{3}{c}{3D Box@0.7} \\ \cline{4-9}
			\rowcolor{mygray}
			\multirow{-2}{*}{Learning Paradigm}&\multirow{-2}{*}{Detector~~~~}	& \multirow{-2}{*}{Modality}&Easy &Moderate &Hard &Easy &Moderate &Hard \\  \hline \hline
			\multicolumn{9}{c}{{\footnotesize Trained with the whole KITTI \texttt{train} set: {\color{red} $3,712$ precisely annotated scenes with $15,654$ vehicle instances}}}\\
			\hline
			\multirow{7}{*}{\textit{Fully supervised}}
			&PIXOR\!~\cite{Yang_2018_CVPR}&LiDAR &87.25 &81.92 &76.01&-&-&-\\
			&VoxelNet\!~\cite{Zhou_2018_CVPR} &LiDAR &89.35 &79.26 &77.39 &77.47 &65.11 &57.73 \\
			&SECOND\!~\cite{Yan2018}&LiDAR &88.07 &79.37 &77.95 &83.13 &73.66 &66.20\\
			&PointRCNN\!~\cite{Shi_2019_CVPR}&LiDAR &89.47 &85.68 &79.10  &85.94 &75.76 &68.32\\
			&PointPillars\!~\cite{Lang_2019_CVPR}&LiDAR &88.35 &86.10 &79.83&79.05 &74.99 &68.30\\
			&Fast PointR-CNN\!~\cite{Chen_2019_ICCV}&LiDAR &88.03 &86.10 &78.17  &84.28 &75.73 &67.39  \\
			&STD\!~\cite{Yang_2019_ICCV}&LiDAR &94.74&89.19&86.42&87.95&79.71&75.09\\
			\hline\hline
			\multicolumn{9}{c}{Trained with a part of KITTI \texttt{train} set: \color{red}$500$ weakly labeled scenes + $534$ precisely annotated instances}\\
			\hline
			\textit{Weakly supervised}& \textbf{Ours} &LiDAR
			&90.11&84.02&76.97&80.15&69.64&63.71\\
			\hline
		\end{tabular}
	}
\end{table}

\subsection{Quantitative and Qualitative Performance}
\label{sec:ex-qp}
\noindent\textbf{Quantitative Results on KITTI \texttt{val} Set (Car).}
In Table\!~\ref{table:KITTIval}, we compare our method
with several leading methods, which all use fully-labeled training data (\ie, $3,712$ precisely-labeled scenes with $15,654$ vehicle instances). However, despite using far less, weakly labeled data, our method yields comparable performance. In addition, as$_{\!}$ there$_{\!}$ is$_{\!}$ no$_{\!}$ other$_{\!}$ weakly$_{\!}$ supervised$_{\!}$ baseline, we$_{\!}$ re-train$_{\!}$ two$_{\!}$ outstanding$_{\!}$ detectors,$_{\!}$ PointRCNN\!~\cite{Shi_2019_CVPR}$_{\!}$  and$_{\!}$  PointPillars\!~\cite{Lang_2019_CVPR},$_{\!}$  under$_{\!}$  two$_{\!}$ relatively$_{\!}$  comparable$_{\!}$ settings, \ie, using \textbf{(i)} 500 precisely labeled scenes (containing 2,176 well-annotated vehicle instances); and \textbf{(ii)} 125 precisely labeled scenes (containing 550 well-annotated instances). This helps further assess the efficacy of our method.
Note that, for training our method, we use 500 scenes with center-click labels and 534 precisely-annotated instances. We find that our method significantly outperforms re-trained PointRCNN and PointPillars, using whether the same amount of well-annotated scenes (500; setting \textbf{(i)}) or the similar number of well-annotated instances (534; setting \textbf{(ii)}).

\noindent\textbf{Quantitative Results on KITTI \texttt{test} Set (Car).} We also evaluate our algorithm on KITTI \texttt{test} set, by submitting our results to the official evaluation server. As shown in Table\!~\ref{table:KITTItest}, though current top-performing methods use much stronger supervision, our method still gets competitive performance against some of them, such as PIXOR\!~\cite{Yang_2018_CVPR} and VoxelNet\!~\cite{Zhou_2018_CVPR}. We can also observe that  there is still room for improvement between weakly- and strong-supervised methods.

\begin{figure}[t]
	\centering
	\includegraphics[width=0.99 \linewidth]{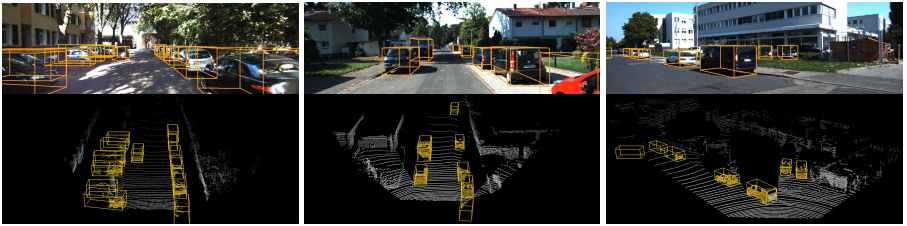}
	\caption{{\textbf{Qualitative results of 3D object detection} (\textbf{Car}) on KITTI \texttt{val} set (\S\ref{sec:ex-qp}). Detected 3D bounding boxes are shown in yellow; images are used only for visualization.  }}
	\label{fig:vresults}
\end{figure}

\begin{table}[t]
	\caption{{\textbf{Evaluation results on KITTI \texttt{val} set (Pedestrian).} See \S\ref{sec:ex-qp} for details.}}
	\centering\small
	\label{table:KITTIvalped}
	\resizebox{0.99\textwidth}{!}{
		\setlength\tabcolsep{3pt}
		\renewcommand\arraystretch{1.08}
		\begin{tabular}{c|r||c|ccc|ccc}
			\hline\thickhline
			\rowcolor{mygray}
			&	 &  	 & \multicolumn{3}{c|}{BEV@0.5} &  \multicolumn{3}{c}{3D Box@0.5} \\ \cline{4-9}
			\rowcolor{mygray}
			\multirow{-2}{*}{Learning Paradigm} &\multirow{-2}{*}{Detector~~~~}	& \multirow{-2}{*}{Modality}&Easy &Moderate &Hard &Easy &Moderate &Hard \\  \hline \hline
			\multicolumn{9}{c}{\footnotesize{Trained with:} \footnotesize{{\color{red}$951$ precisely labeled scenes with $2,257$ pedestrian instances}}}\\
			\hline
			\multirow{5}{*}{\textit{Fully supervised}}
			&PointPillars\!~\cite{Lang_2019_CVPR} &LiDAR &71.97 &67.84 &62.41 &66.73 &61.06 &56.50\\
			&PointRCNN\!~\cite{Shi_2019_CVPR} &LiDAR &68.89&63.54&57.63&63.70&69.43&58.13\\
			&Part-$A^2$\!~\cite{Shi_2019_part} &LiDAR &-&-&-&70.73&64.13&57.45\\
			&VoxelNet\!~\cite{Zhou_2018_CVPR}&LiDAR &70.76&62.73&55.05&-&-&-\\
			&STD\!~\cite{Yang_2019_ICCV}&LiDAR &75.90&69.90&66.00&73.90&66.60&62.90\\
			\hline
			\hline
			\multicolumn{9}{c}{\footnotesize{Trained with:} \footnotesize{{\color{red}$951$ weakly labeled scenes with $515$ pedestrian instances}}}\\
			\hline
			\textit{Weakly supervised}&\textbf{Ours} &LiDAR &74.79&70.17&66.75&74.65&69.96&66.49\\
			\hline	
		\end{tabular}
	}
\end{table}

\noindent\textbf{Qualitative Results.} Fig.\!~\ref{fig:vresults} depicts visual results of a few representative scenes from KITTI \texttt{val} set, showing that our model is able to produce high-quality 3D detections of vehicles that are highly occluded or far away from the ego-vehicle.

\noindent\textbf{Quantitative Results on KITTI \texttt{val} Set (Pedestrian).}  We also report~our performance on \textbf{Pedestrian} class (with  0.5 IoU threshold). In this case, our~model is trained with 951 click-labeled scenes and 25\% (515) of precisely annotated pedestrian instances. More training details can be found in the supplementary material. As shown in Fig.\!~\ref{table:KITTIvalped}, our method shows very promising results, demonstrating its good generalizability and advantages when using less supervision.

\subsection{Diagnostic Experiment}
\label{sec:ablation}

As the ground-truth for KITTI \texttt{test} set is not available and the access to the test server is limited, ablation studies are performed over the \texttt{val} set (see Table\!~\ref{table:abla}).

\begin{table}[t]
	\centering
	\caption{{\textbf{Ablation study on  KITTI \texttt{val} set (Car).} See \S\ref{sec:ablation} for details.}}
	\label{table:abla}
	\resizebox{0.99\textwidth}{!}{
		\setlength\tabcolsep{1pt}
		\renewcommand\arraystretch{1.08}
		\begin{tabular}{c|r||c|ccc|ccc}
			\hline\thickhline
			\rowcolor{mygray}
			\multicolumn{2}{c||}{}  &	& \multicolumn{3}{c|}{BEV@0.7} &  \multicolumn{3}{c}{3D Box@0.7} \\ \cline{4-9}
			\rowcolor{mygray}
			\multicolumn{2}{c||}{\multirow{-2}{*}{Aspects~~~~}}	& \multirow{-2}{*}{Training Setting}&Easy &Moderate &Hard&Easy &Moderate &Hard \\  \hline \hline
			\multicolumn{2}{c||}{\textbf{Full model}} &{\tabincell{c}{500 weakly labeled scenes \\+$25\%$ (534) precisely annotated instances}}
			&88.56  &84.99  &84.74  &84.04  &75.10  &73.29\\
			\hline\hline
			\multicolumn{2}{c||}{\tabincell{c}{More precisely \\annotated BEV maps}}&\tabincell{c}{500 weakly labeled scenes \\+$25\%$ (534) more precisely annotated instances}
			&88.52 &84.57 &85.02 &85.67 &75.13 &73.92\\
			\hline\hline
			\multicolumn{2}{c||}{}&\tabincell{c}{3,712 weakly labeled scenes \\+534 precisely annotated instances}
			&88.81&86.98&85.76&86.08&76.04&74.97\\
			\cline{3-9}
			\multicolumn{2}{c||}{}&\tabincell{c}{1,852 weakly labeled scenes \\+$25\%$ precisely annotated instances}
			&89.11 &85.95 &85.52 &87.14 &76.78 &76.56\\
			\cline{3-9}
			\multicolumn{2}{c||}{\multirow{-5}{*}{More training data}}&\tabincell{c}{3,712 weakly labeled scenes \\+$25\%$ precisely annotated instances}
			&89.32&86.17&86.31&87.57&77.62&76.94\\
			\hline
		\end{tabular}
	}
\end{table}

\noindent\textbf{Robustness to Inaccurate BEV Annotations.} As discussed in \S\ref{sec:3Dannotation}, the annotations over the BEV maps are weak and inaccurate. To examine our robustness to inaccurate BEV annotations, we retrain our model with precise BEV annotations inferred from groundtruth 3D annotations. From Table~\ref{table:abla}, only marginal improvements are observed, verifying our robustness to noisy BEV annotations.

\noindent\textbf{More Training Data.} To demonstrate the potential of our weakly supervised 3D object detection scheme, we probe the upper bound by training on additional data.
As evidenced by the results in the last three rows in Table~\ref{table:abla}, with the use of more training data, gradual performance boosts can indeed be achieved.

\subsection{Performance as An Annotation Tool}
\label{sec:ex-an}

Our model, once trained, can be used as an 3D object annotator. It only consumes part of KITTI \texttt{train} set, allowing us to explore its potential for assisting annotation. Due to its specific network architecture and click-annotation guided learning paradigm, it supports both automatic and active annotation modes.

\noindent\textbf{Automatic Annotation Mode.} For a given scene, it is straightforward to use our predictions as pseudo annotations, resulting in an automatic working mode. In such a setting, our method takes around 0.1 s for per car instance annotation. Previous 3D detection methods can also work as automatic annotators in this way. However, as they are typically trained with the whole KITTI \texttt{train} set, it is hard to examine their annotation quality.

\noindent\textbf{Active Annotation Mode.} In the active mode, human annotators first click on object centers in BEV maps, following the labeling strategy detailed in \S\ref{sec:3Dannotation}. Then, for each annotated center, 25 points are uniformly sampled from the surrounding 0.4 m$\times$0.4 m region (0.1 m interval). These points are used as the centers of cylindrical proposals and the foreground masks around them are generated according to Eq.~\ref{eq:Pseudo}. Then, we use our Stage-2 model to predict the cuboids, from which the one with largest confidence score is selected as the final annotation. About 2.6 s is needed for annotating each car instance in our active annotation mode, whereas humans take 2.5 s for center-click annotation on average.

\begin{table}[t]
	\caption{{\textbf{Comparison of annotation quality on KITTI \texttt{val} set} (see \S\ref{sec:ex-an}). }}
	\centering
	\label{table:ex-an-KITTIval}
	\resizebox{0.99\textwidth}{!}{
		\setlength\tabcolsep{5pt}
		\renewcommand\arraystretch{1.08}
		\begin{tabular}{c|c||c|c|ccc|ccc}
			\hline\thickhline
			\rowcolor{mygray}
			&&&Speed&\multicolumn{3}{c|}{BEV@0.5} &  \multicolumn{3}{c}{3D Box@0.5} \\
			\cline{5-10}
			\rowcolor{mygray}
			\multirow{-2}{*}{\!\!Learning Paradigm\!\!}&\multirow{-2}{*}{\!\!Method\!\!} &\multirow{-2}{*}{Mode} &(sec./inst.)  &Easy &\!\!Moderate\!\! &Hard &Easy&\!\!Moderate\!\! &Hard \\
			\hline \hline
			\multicolumn{10}{c}{Trained with the whole KITTI \texttt{train} set: \color{red}3,712 well-labeled scenes with 15,654 vehicle instances}\\
			\hline
			\textit{Fully Supervised}&\cite{lee2018leveraging} &Active&3.8 &-&-&-&-&-&88.33\\\hline \hline
			\multicolumn{10}{c}{Trained with KITTI \texttt{train}+\texttt{val}: {\color{red}$7,481$ scenes} (implicitly using 2D instance segmentation annotations)}\\
			\hline
			\textit{Fully-Supervised}&\cite{zakharov2019autolabeling}&Auto&8.0 &80.70 &63.36&52.47&63.39&44.79&37.47\\
			\hline \hline
			\multicolumn{10}{c}{Trained with a part of KITTI \texttt{train} set: \color{red}500 weakly labeled scenes + 534 precisely annotated instances}\\ \hline
			\multirow{2}{*}{\!\!\textit{Weakly Supervised}\!\!}&\multirow{2}{*}{\textbf{Ours}}
			&Auto&0.1&96.33 &89.01 &88.52 &95.85 &89.14 &88.32 \\
			&&Active&2.6&99.99 &99.92 &99.90 &99.87 &90.78 &90.14\\
			\hline	
		\end{tabular}
	}
\end{table}

\begin{table}[!t]
	\centering
	\caption{{\textbf{$_{\!}$Performance$_{\!}$ of$_{\!}$ PointRCNN\!~\cite{Shi_2019_CVPR}$_{\!}$ and$_{\!}$ PointPillars\!~\cite{Lang_2019_CVPR}$_{\!}$ when$_{\!}$ trained$_{\!}$ using$_{\!}$ different$_{\!}$ annotations$_{\!}$ sources.} Results$_{\!}$ are$_{\!}$ reported$_{\!}$ on$_{\!}$  KITTI$_{\!}$ \texttt{val}$_{\!}$ set (\S\ref{sec:ex-an}).}}
	\label{table:ex-an-KITTIval2}
	\resizebox{0.99\textwidth}{!}{
		\setlength\tabcolsep{6pt}
		\renewcommand\arraystretch{1}
		\begin{tabular}{r||c|ccc|ccc}
			\hline\thickhline
			\rowcolor{mygray}
			&  	& \multicolumn{3}{c|}{BEV@0.7} &  \multicolumn{3}{c}{3D Box@0.7} \\ \cline{3-8}
			\rowcolor{mygray}
			\multirow{-2}{*}{Detector~~~~}	& \multirow{-2}{*}{Annotation Source}&Easy &Moderate &Hard&Easy &Moderate &Hard \\  \hline \hline
			\multirow{3}{*}{PointRCNN\!~\cite{Shi_2019_CVPR}}
			&Manual &90.21 &87.89 &85.51 &88.45&77.67&76.30\\
			&Automatic (ours)&88.02 &85.75 &84.27 &83.22 &74.54 &73.29\\
			&Active (ours)&88.64 &85.41 &84.94 &84.21 &76.08 &74.91\\
			\hline\hline
			\multirow{3}{*}{PointPillars\!~\cite{Lang_2019_CVPR}}
			&Manual&89.64&86.46&84.22&85.31&76.07&69.76\\
			&Automatic (ours)&88.55 &85.62 &83.84 &84.79 &74.18 &68.52\\
			&Active (ours)&88.94 &85.88 &83.86 &84.53 &75.03 &68.63\\
			\hline
		\end{tabular}
	}
\end{table}

\noindent\textbf{Annotation Quality.} Table~\ref{table:ex-an-KITTIval}  reports the evaluation results for our annotation quality on KITTI \texttt{val} set. Two previous annotation methods\!~\cite{zakharov2019autolabeling,lee2018leveraging} are included. \cite{lee2018leveraging} is a fully supervised deep learning annotator, trained with the whole KITTI \texttt{train} set. It only works as an active model, where humans are required to provide object anchor clicks. \cite{zakharov2019autolabeling} requires synthetic data for training and relies on MASK-RCNN\!~\cite{he2017mask}, so it implicitly uses 2D instance segmentation annotations. It works in an automatic mode.
The scores for these models are borrowed from the literature, as their implementations are not released. Following their settings\!~\cite{zakharov2019autolabeling,lee2018leveraging}, scores with 0.5 3D IoU criterion are reported.  As seen, our model produces high-quality annotations, especially in the active mode. Our annotations are more accurate than\!~\cite{zakharov2019autolabeling,lee2018leveraging}, with much less and weak supervision. Considering our fast annotation speed ($0.1$$-$$2.6$ s per instance), the results are very significant.

\noindent\textbf{Suitability for 3D Object Detection.} To investigate the suitability of our labels for 3D object detection, we use our re-labeled KITTI \texttt{train} set (\textbf{Car}) to re-train PointPillars\!~\cite{Lang_2019_CVPR} and PointRCNN\!~\cite{Shi_2019_CVPR}, which show leading performance with released implementations. During training, we use their original settings. From Table\!~\ref{table:ex-an-KITTIval2}, we can observe that the two methods only suffer from small performance drops when using our labels.

\section{Conclusion and Discussion}

This work has made an early attempt to train a 3D object detector using limited and weak supervision. In addition,  our detector can be extended as an annotation tool, whose performance was fully examined in both automatic and active modes. Extensive experiments on KITTI dataset demonstrate our impressive results, but also illustrate that there is still room for improvement. Given the massive number of algorithmic breakthroughs over the past few years, we can expect a flurry of innovation towards this promising direction.

\def\thesection{\Alph{section}}
\makeatletter
\renewcommand{\@maketitle}{
\newpage
 \null
 \vskip 2em%
 \begin{center}%
  {\LARGE \@title \par}%
 \end{center}%
 \par} \makeatother

\pagestyle{headings}
\mainmatter
\def\ECCVSubNumber{1956}

\title{Weakly Supervised 3D Object Detection from \\Lidar Point Cloud \\\textit{Supplementary Material}}

\titlerunning{Weakly Supervised 3D Object Detection}
\authorrunning{Q. Meng, W. Wang, T. Zhou, J. Shen, L. Van Gool, D. Dai}

\author{Qinghao Meng$^{1}$\and
	\Letter Wenguan Wang$^{2}$\and
	Tianfei Zhou$^{3}$\and\\
	Jianbing Shen$^{3}$\and
	Luc Van Gool$^{2}$\and
	Dengxin Dai$^{2}$}
\institute{$^{1}$School of Computer Science, Beijing Institute of Technology\\$^{2}$ETH Zurich~~~~$^{3}$Inception Institute of Artificial Intelligence\\
	\url{https://github.com/hlesmqh/WS3D}
}

\maketitle

In this document, we first give more implementation details of applying our model for 3D pedestrian detection (see \S\ref{sec: Strategy}). Later, in \S\ref{sec:Visualization}, we give some visual results for 3D pedestrian detection on KITTI~\!\cite{KITTI} \texttt{val} set. Then, in \S\ref{sec:annotator}, we compare the annotation results by applying our trained model as annotation tools, working in two annotation modes, \ie, automatic and active. Finally, we discuss some representative failure cases in \S\ref{sec:failure}.

\section{Weakly Supervised 3D Pedestrian Detection}
\label{sec: Strategy}

We specify some modifications for adapting our method for \textbf{Pedestrian} class.

\noindent\textbf{Data Preparation.} For the KITTI training set which contains a total of $3,712$ scenes, there are only
$951$ scenes contain pedestrian labels. Considering the small amount of training samples, we use the weakly annotated BEV maps of the $951$ scenes to train our Stage-1 model. We randomly choose $515$, nearly $25\%$ in $2,257$ samples in those scenes as the training data for our Stage-2 model. Compared with prior fully-supervised algorithms which leverage all the exhaustively annotated $951$ scenes with $2,257$ pedestrian samples, we use far less and weak supervision. To reduce futile false negative responses and speeding up the CA-NMS process for better effectiveness, following~\!\cite{Lang_2019_CVPR}, we set the $x,z$ range of the searching region for pedestrian as $[(-20,20), (0,48)]$, respectively.

\noindent\textbf{Pseudo Foreground Groundtruth Generation.} For \textbf{Car} class, we use an \textit{ellipsoid-shaped} 3D Gaussian distribution for pseudo \textit{soft} foreground groundtruth generation (see Eq. {\color{red}1}). For \textbf{Pedestrian} class, we instead  directly use a \textit{pillar} (cylinder) to generate pseudo \textit{binary} masks. This is because, compared with vehicles which are typically  presented as elongated rectangles on BEV maps, the shapes of human on the BEV maps are more like regular squares. The radius of the pillars are uniformly set as 0.4 m.

\noindent\textbf{Cylindrical 3D Proposal Generation.} Considering the small size of pedestrians, we generate the cylindrical proposal with a 1 m radius over ($x$, $z$)-plane (4 m radius for vehicle). For each groundtruth, the proposals whose center-distances to it are less than $0.5$ m are selected as its training samples.

\noindent\textbf{Training}. For \textbf{Car} class, we use Adam optimizer with an initial learning rate 0.002 and weight decay 0.0001. In Stage-1, we train the network for 8K iterations with batch-size
25. In Stage-2, the whole training process takes 50K iterations with batch-size 800. For \textbf{Pedestrian} class, we use the same parameters to train Stage-1 model and reduce the training process to 20K iterations in Stage-2.


\begin{figure}[t]
	\centering
	\includegraphics[width=0.99 \linewidth]{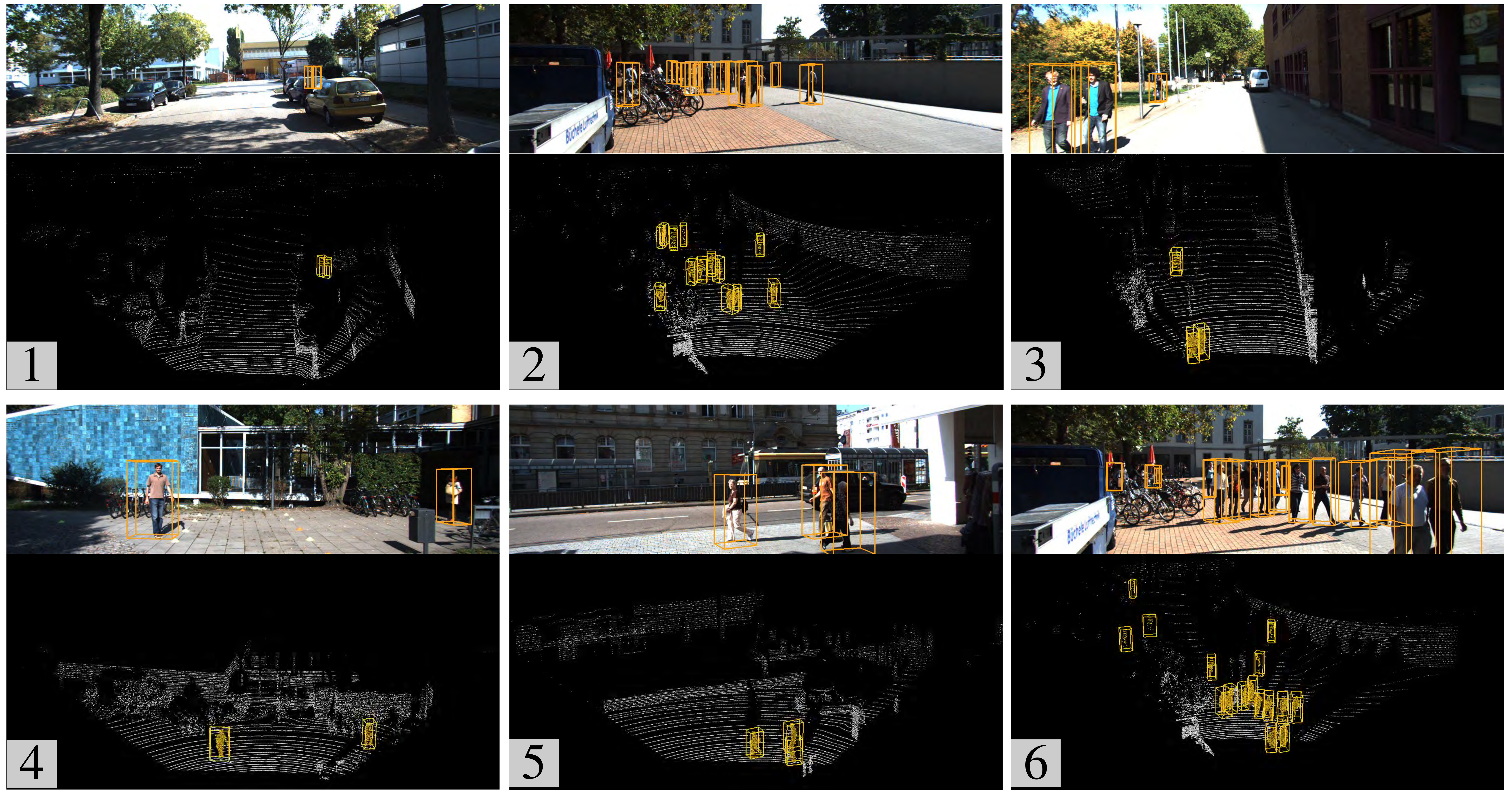}
	\caption{{\textbf{Qualitative results of 3D object detection} (Pedestrian) on KITTI \texttt{val} set. Detected 3D bounding boxes on image and point cloud pairs are depicted in yellow. }}
	\label{fig:vresults}
\end{figure}

\section{Qualitative Results on KITTI \texttt{val} Set (Pedestrian)}
\label{sec:Visualization}

In Fig.~\!\ref{fig:vresults}, we visualize representative outputs of our model on KITTI \texttt{val} set for \textbf{Pedestrian} class. As seen, for simple cases of non-occluded objects in reasonable distance which we got enough number of points, our model outputs remarkably accurate 3D bounding boxes (like subfigures 3, 4 and 5). Second, we are surprised to find that our model can even correctly predict some highly occluded ones (subfigure 1) and works well in several crowded scenes (subfigures 2 and 6). This proves that our proposed detector not only handles well vehicles, but also adapts to other challenging classes in autonomous driving scenes, under less and easily acquired supervision.

\begin{figure}[t]
	\centering
	\includegraphics[width=0.9 \linewidth]{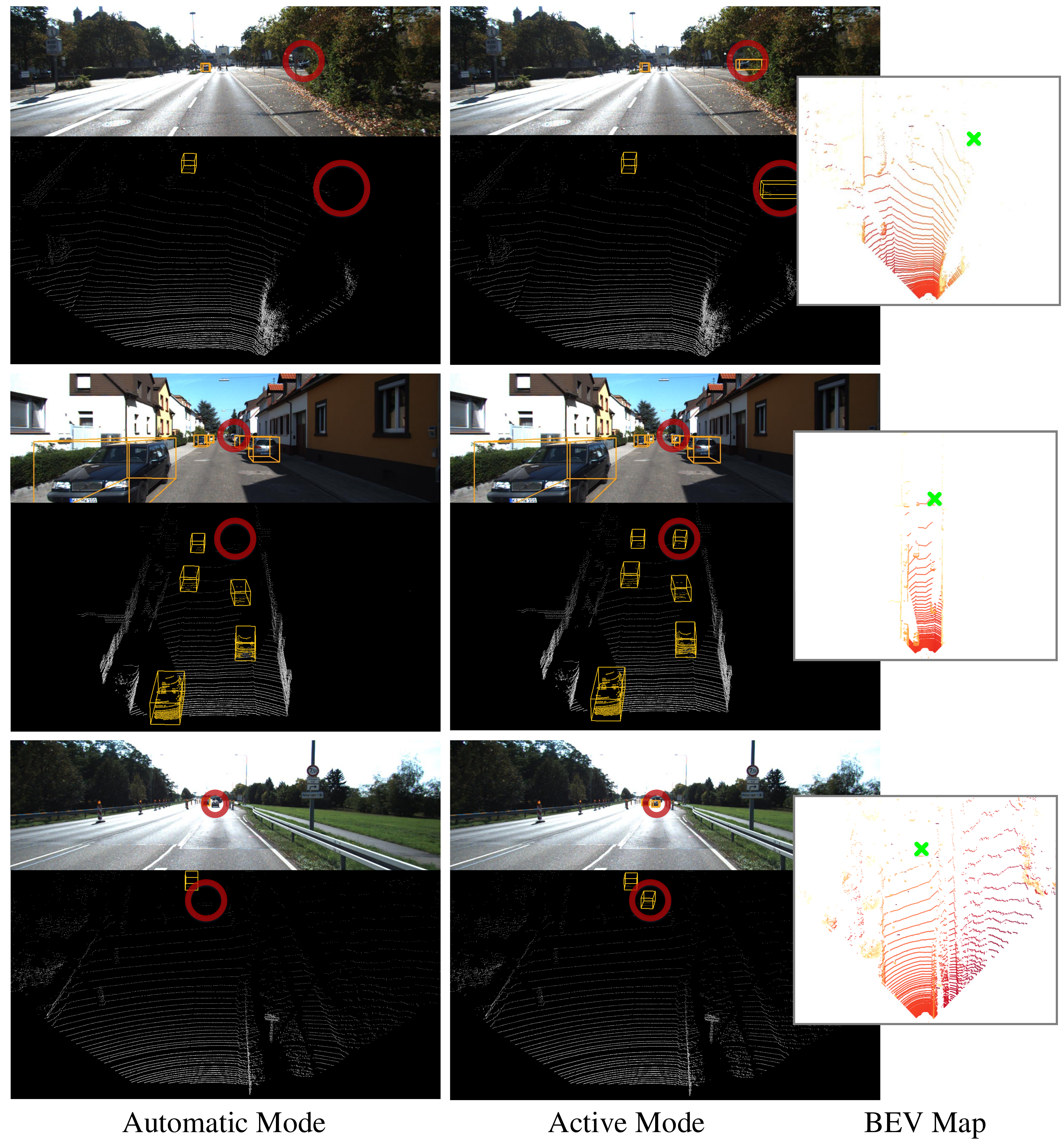}
	\caption{{\textbf{Annotation results for 3D object detection} (Car) on KITTI \texttt{val} set. labeled 3D bounding boxes on image and point cloud pairs are depicted in yellow. The improved annotations are highlighted by red circles. Zoom-in for details.}}
	\label{fig:annresults}
\end{figure}

\section{Annotation Results on KITTI \texttt{val} Set (Car)}
\label{sec:annotator}

Due to our specific network architecture and weakly supervised learning protocol, our model, once trained, can be applied as an annotation tool, which allows automatic and active annotation modes, to improve annotation efficiency. In Fig.~\!\ref{fig:annresults}, we present some annotation results generated from automatic and active modes. It can be observed that in most cases our model with automatic mode can obtain high-quality annotation results. In addition, our model allows human annotators to place extra clicks on the centers of desired objects, thus the inferior or missing predictions can be corrected. In the active mode, with the weak supervision provided by human annotators, better proposals can be generated around the click points and thus leading to improved predictions.

\section{Failure Cases on KITTI \texttt{val} Set (Car\&Pedestrian)}
\label{sec:failure}

\begin{figure}[t]
	\centering
	\includegraphics[width=0.99 \linewidth]{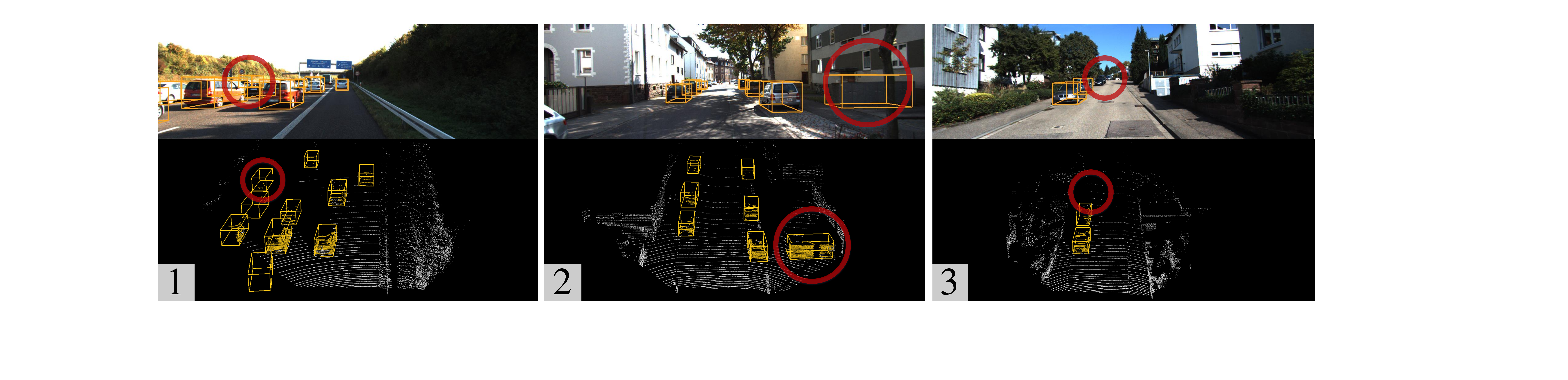}
	\caption{{\textbf{Failure cases of 3D car detection} on KITTI \texttt{val} set. Predicted 3D bounding boxes on image and point cloud pairs are depicted in yellow.  The inaccurate predictions are highlighted by red circles. Zoom-in for details.}}
	\label{fig:fcases_car}
\end{figure}

\begin{figure}[t]
	\centering
	\includegraphics[width=0.99 \linewidth]{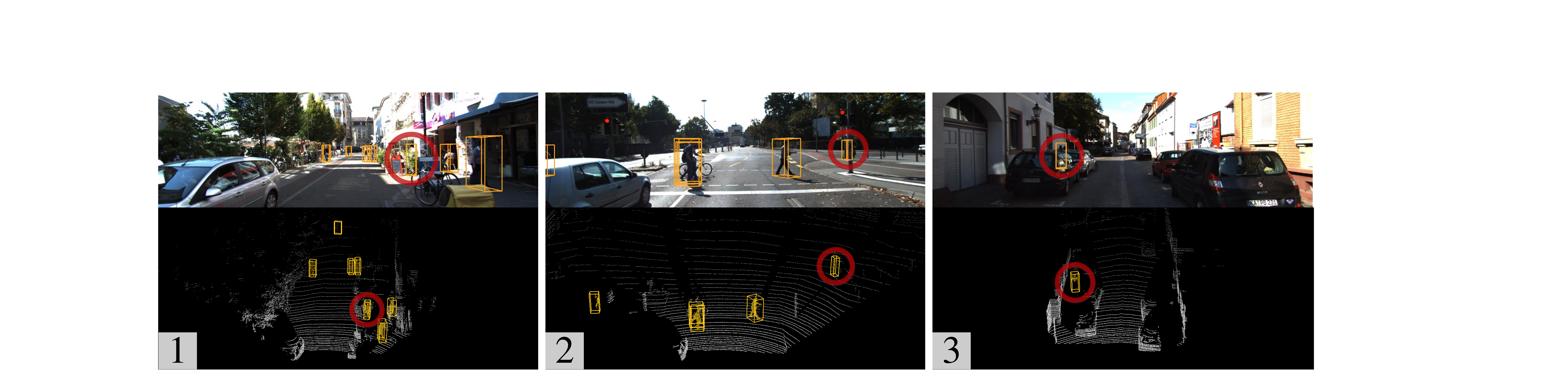}
	\caption{{\textbf{Failure cases of 3D pedestrian detection} on KITTI \texttt{val} set. Predicted 3D bounding boxes on image and point cloud pairs are depicted in yellow. The inaccurate predictions are highlighted by red circles. Zoom-in for details.}}
	\label{fig:fcases_ped}
\end{figure}

Though our predictions for cars are particularly accurate, there are still common failure modes, summarized in Fig.~\!\ref{fig:fcases_car}. The \textit{first} type of common mistakes are caused by the heavy occlusions, such as the vehicle in subfigure 1, highlighted by the red circle, is predicted with wrong height. We think leveraging more contextual information may be helpful. The \textit{second} type of challenge is caused by some background objects, like the large box in subfigure 2, which has a similar shape of vehicle. Our model is easily confused, as these background objects look very like vehicles in the point cloud. \textit{Third}, for some challenging cases where the foreground points are extremely sparse, our model is hard to make accurate predictions. Subfigure 3 shows a typical example, where the points of the highlighted vehicles are very few due to the occlusion of hillside. The last two problem can be partially mitigated by considering extra appearance information from  camera images. Detecting pedestrians is more challenging and leads to similar lapses. As we can see in Fig.~\!\ref{fig:fcases_ped}, the model is occasionally confused by cylindrical obstacles such as the plant in subfigure 1 and the pole in subfigure 2, which are false positives. In subfigure 3, the two pedestrians are very close and highly occluded, making our output mix them together. Above challenges also indicate possible directions for our future efforts.

\end{document}